\pdfoutput=1

\documentclass[11pt]{article}

\usepackage[preprint]{acl}

\usepackage{times}
\usepackage{latexsym}

\usepackage[T1]{fontenc}

\usepackage[utf8]{inputenc}

\usepackage{microtype}

\usepackage{inconsolata}

\usepackage{graphicx}
\usepackage{times}
\usepackage{bbm}
\usepackage{latexsym}
\usepackage[utf8]{inputenc}
\usepackage{subcaption}
\usepackage{graphicx}
\usepackage{booktabs}
\usepackage{fontawesome5}
\usepackage[misc]{ifsym}
\usepackage{microtype}
\usepackage{amsmath}
\usepackage{mathtools}
\usepackage[T1]{fontenc}

\usepackage[utf8]{inputenc}
\usepackage{subcaption}
\usepackage{graphicx}
\usepackage{booktabs}
\usepackage{fontawesome5}
\usepackage[misc]{ifsym}
\usepackage{hyperref}
\usepackage{url}
\usepackage[utf8]{inputenc} 
\usepackage[T1]{fontenc}    
\usepackage{hyperref}       
\usepackage{url}            
\usepackage{booktabs}       
\usepackage{amsfonts}       
\usepackage{nicefrac}       
\usepackage{microtype}      
\usepackage{graphicx}
\usepackage{natbib}
\usepackage{doi}
\usepackage{xcolor}         
\usepackage{amsmath}
\usepackage[linesnumbered,ruled,vlined]{algorithm2e}
\usepackage{multirow}
\usepackage{adjustbox}
\SetKwInput{kwInit}{Init}
\SetKwInput{KwRequire}{Require}
\usepackage{graphicx, caption, subcaption}

\title{S2MoE: Robust Sparse Mixture of Experts via Stochastic Learning}

\author{
    Giang Do\thanks{Corresponding author} \quad Hung Le \quad Truyen Tran  \\
    Applied Artificial Intelligence Institute (A2I2), Deakin University \\
    \texttt{\{s224363215,thai.le,truyen.tran\}@deakin.edu.au}\\
  }

\begin{document}
\maketitle
\begin{abstract}
Sparse Mixture of Experts (SMoE) enables efficient training of large language models by routing input tokens to a select number of experts. However, training SMoE remains challenging due to the issue of representation collapse. Recent studies have focused on improving the router to mitigate this problem, but existing approaches face two key limitations: (1) expert embeddings are significantly smaller than the model’s dimension, contributing to representation collapse, and (2) routing each input to the Top-K experts can cause them to learn overly similar features. In this work, we propose a novel approach called Robust Sparse Mixture of Experts via Stochastic Learning (S2MoE), which is a mixture of experts designed to learn from both deterministic and non-deterministic inputs via Learning under Uncertainty. Extensive experiments across various tasks demonstrate that S2MoE achieves performance comparable to other routing methods while reducing computational inference costs by \textbf{28\%}.
\end{abstract}

\section{Introduction}
Sparse Mixture of Experts (SMoE) models have achieved notable success in natural language processing (NLP) and visual representation learning tasks \citep{du_glam_2022, fedus_switch_2022, riquelme2021scalingvisionsparsemixture, shen-etal-2023-scaling}. These advancements build on the Transformer architecture \citep{NIPS2017_3f5ee243} and its variants \citep{child2019generating, dai2019transformerxl}, which leverage large datasets and significant compute resources. However, training large Transformer models can be prohibitively expensive, requiring extensive compute hours \citep{kaddour2023challenges}. To address this, SMoE models activate only a subset of experts for each input, reducing inference time compared to dense models \citep{shazeer2017outrageously, zoph2022stmoe, artetxe2022efficient, krajewski2024scaling}.

\begin{figure}[t]
    \centering
    \begin{center}
        \centerline{\includegraphics[width=0.5\textwidth]{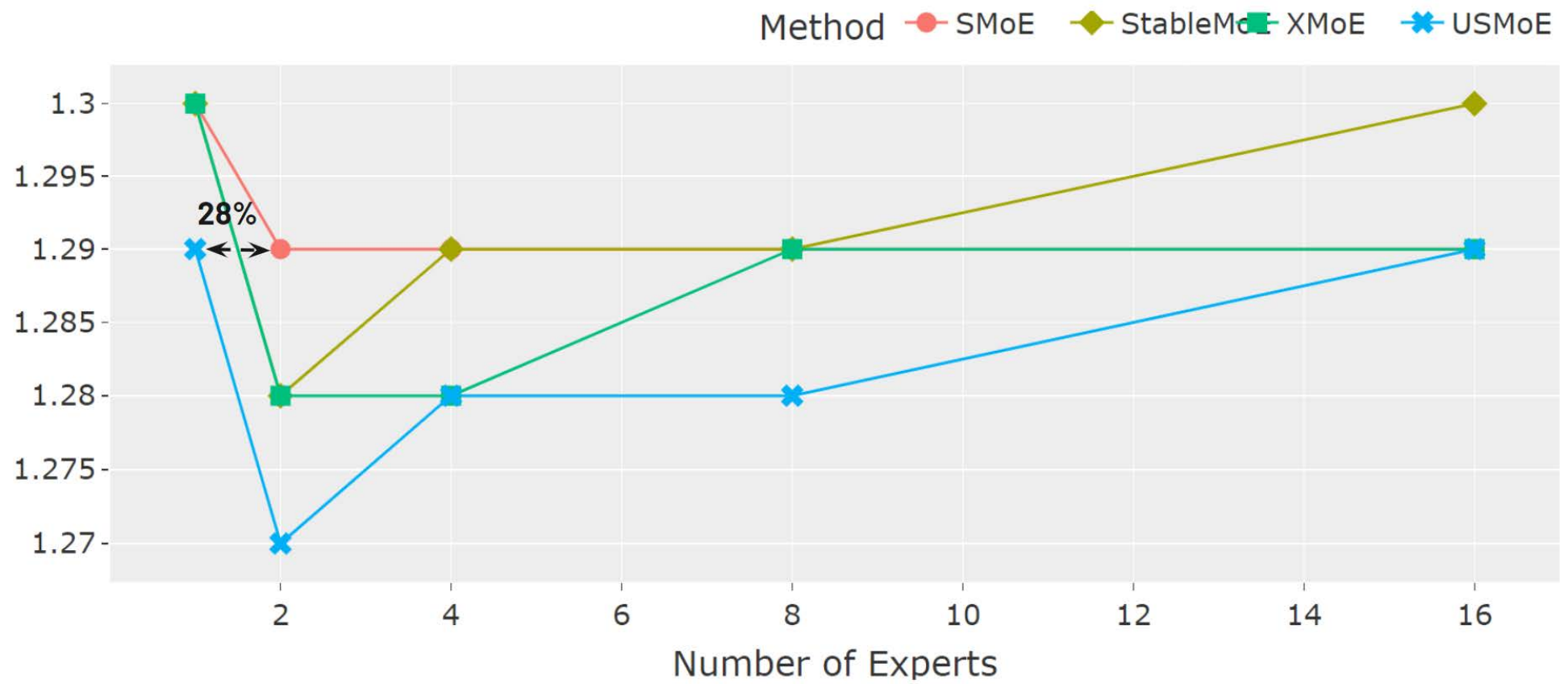}}	
        \vspace*{-0.1in}
	\caption{BPC (\textit{Bits-per-character}) on the \texttt{Text8} dataset with varying numbers of experts used for inference. S2MoE requires the activation of only one expert to achieve comparable performance with other routing methods, resulting in a savings of \textbf{28\%} in computational inference costs. All methods have the same FLOPs. }
	\label{fig:Flops}
    \end{center}
    \vskip -0.3in
\end{figure}

\noindent Despite promising results, SMoE models face the challenge of \textit{representation collapse}, where either a few experts dominate the routing or all experts learn similar representations \cite{chi_representation_2022, chen_towards_2022}. To address this, research has focused on improving router policies \cite{chi2022representation, chen2023sparse, do2023hyperrouter}. One solution, SMoE-Dropout \cite{chen_sparse_2023}, freezes a randomly initialized router throughout training and gradually increases the number of active experts. However, these existing approaches have two limitations: (1) the expert embeddings are much smaller than the model dimension, leading to representation collapse, and (2) routing each input to the Top-K experts can cause them to learn similar features.

\noindent To address these limitations, this work proposes a novel approach called Robust Sparse Mixture of Experts via Stochastic Learning (S2MoE) to enhance expert knowledge and prevent overlap in their learning. Instead of feeding the same input to the Top-K experts, S2MoE utilizes a Gaussian noise to enhance feature learning prior to expert selection, a concept that has been validated in the vision domain~\cite{5570958,1232360,chen2024composedimageretrievaltext}. By doing this, S2MoE can enhance expert learning efficiency during training and reduce the representation collapse issue. To showcase its effectiveness, we perform comprehensive evaluations across various NLP tasks, comparing S2MoE with several state-of-the-art SMoE routing strategies. Additionally, S2MoE reaches the same performance levels with fewer experts during inference, greatly improving the efficiency of deploying LLMs in real-world applications. Figure \ref{fig:Flops} demonstrates that S2MoE requires the activation of only one expert to achieve comparable performance with other routing methods, resulting in a savings of \textbf{28\%} in computational inference costs.

\vspace*{-0.11in}
\section{Related Work}
\vspace*{-0.11in}
{\bf Sparse Mixture of Experts (SMoE). } 

Sparse Mixure of Experts (SMoE) building on the Mixture of Experts framework \citep{jacobs1991,jordan1994}, gained traction with large language models and has since been applied in various fields, including computer vision and speech recognition \citep{NEURIPS2022_2f00ecd7,NEURIPS2021_48237d9f}.  However, SMoE encounters the challenge of representation collapse, where experts produce similar outputs. To combat this, various methods have emerged, such as XMoE, which uses low-dimensional routing scores \citep{chi2022representation}, and SMoE-dropout, which activates more experts gradually \citep{chen2023sparse}. Other strategies include HyperRouter \citep{do2023hyperrouter} and StableMoE \citep{dai2022stablemoe}, both aimed at improving the stability and robustness of routers. Despite these innovations, representation collapse remains a persistent issue \citep{pham2024competesmoe}. Our approach differs by emphasizing enhanced feature learning, which helps expand experts' knowledge and reduce the collapse issue.

\noindent
{\bf Learning under Uncertainty. } Learning under Uncertainty have a long history that consists well-known research topic in Machine Learning such as Dropout~\cite{JMLR:v15:srivastava14a}, Bayesian neural networks ~\cite{article} and noise regularized learning ~\cite{articlenoise}. Some studies \cite{8628917,MAHARANA202291,8462624} have applied learning under uncertainty in data augmentation, a common technique in the vision domain that helps models improve robustness and reduce overfitting. Additionally, ~\cite{chen2024composedimageretrievaltext} enhanced feature learning for vision models by incorporating Gaussian noise generation.

\section{Methodology}

We propose a novel model, the Stochastic Sparse Mixture of Experts (S2MoE), which is a mixture of experts designed to learn from both deterministic and non-deterministic inputs. As illustrated in Figure~\ref{fig:S2MoE}, our method consists of two parts: (1) learning with the original input and (2) learning with noise-added input. To regulate the quality of the noise generation process, we introduce uncertainty loss, as shown in the Equation ~\ref{eq:nloss}.
\vspace*{-0.1in}
\subsection{Preliminaries}

\textbf{Sparse Mixture of Experts.}\; The Sparse Mixture of Experts (SMoE) is typically a transformer architecture that replaces the multi-layer perceptron (MLP) layers in standard transformers with Mixture of Experts (MoE) layers, inspired by \citep{shazeer2017outrageously}. 
Given $\mathbf{x} \in \mathbb{R}^{n \times d}$ as the output of the multi-head attention (MHA) layer, the result of the SMoE with $N$ experts is a weighted sum of each expert's computation, $E_i(x)$, weighted by the router function $\mathcal{S}(x)$:

\begin{align} \label{eq:smoe}
    f_{\mathrm{SMoE}}(\boldsymbol{x})=\sum_{i=1}^N \mathcal{S}(\boldsymbol{x})_i \cdot E_i(\boldsymbol{x})
\end{align}

Where $\mathcal{S}(x)$ is computed by $TopK$ function as below the Equation ~\ref{eq:gate}, and $W_e$ is a learnable experts embeddings.

\begin{equation}\label{eq:gate}
\mathcal{S}(\boldsymbol{x})=\operatorname{TopK}(\operatorname{softmax}(W_e \cdot x), k) 
\end{equation}

\vspace*{-0.3in}
\begin{equation}\label{eq:topk}
\resizebox{.9\hsize}{!}{$
 \operatorname{TopK}(\boldsymbol{v}, k) =
\begin{cases}
\boldsymbol{v_i} & \text{if } \boldsymbol{v_i} \text{ is in the top } k \text{ largest of } \boldsymbol{v}, \\
-\infty & \text{otherwise}.
\end{cases}
$}  \notag
\end{equation}

\subsection{Robust Sparse Mixture of Experts via Stochastic Learning (S2MoE)}
\label{arch}
\begin{figure}[t]
    \begin{center}
        \centerline{\includegraphics[width=0.4\textwidth]{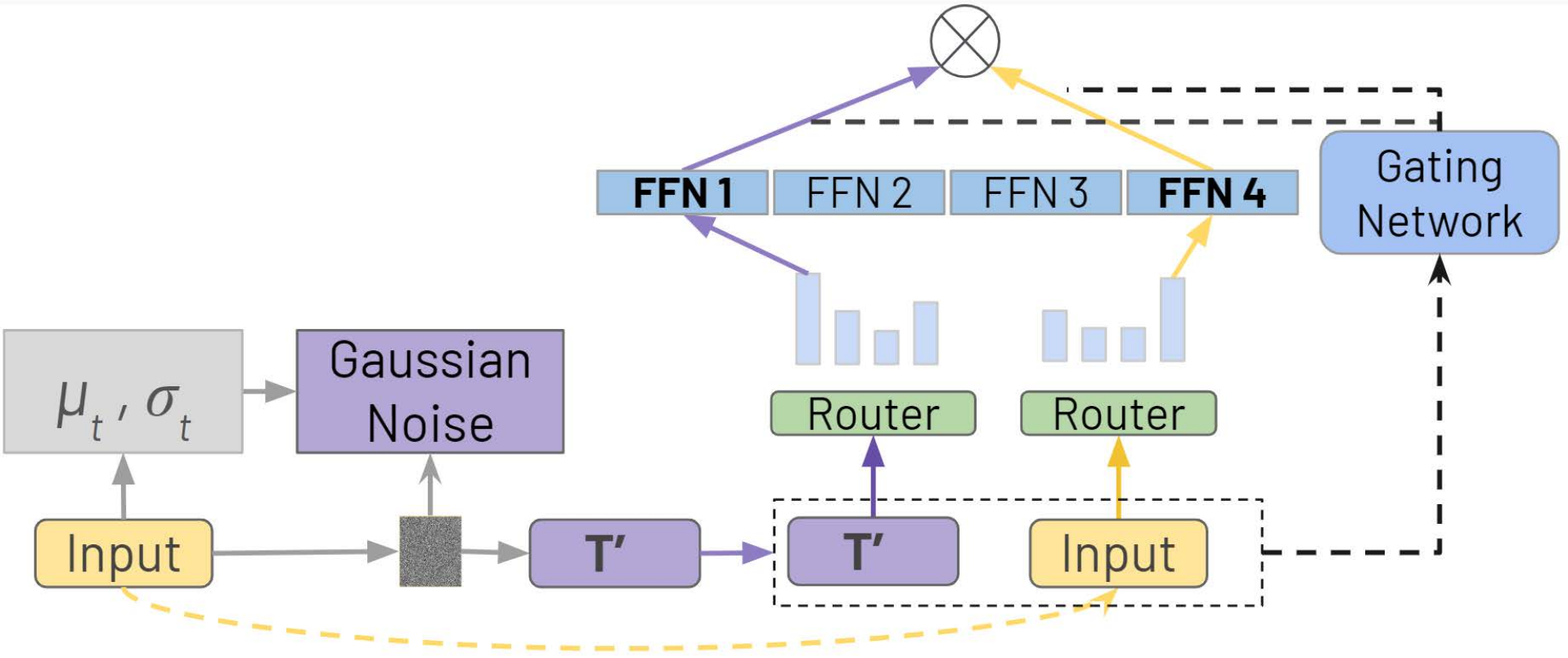}}	
	\caption{An illustration of our S2MoE that enhances model knowledge through Gaussian noise generation. The method involves two components: learning from the original input and the noise-augmented input concurrently through SMoE, with their outputs combined by a gating network implemented as a 1-layer MLP. Best viewed in colors.}
	\label{fig:S2MoE}
    \end{center}
    \vskip -0.4in
\end{figure}

\textbf{Uncertainty Modeling.}\; Inspired by ~\cite{chen2024composedimageretrievaltext}, we introduce the Gaussian Noise Module as Figure~\ref{fig:S2MoE}, which directly applies to the representation space to enhance model feature learning. Given a representation $\mathbf{x} \in \mathbb{R}^{n \times d}$ and $\mu_x$, $\sigma_x$ representing the mean and standard deviation of the Gaussian noise calculated per batch from the feature $x$, the noise-augmented input $\hat{x}$ is defined by the following formula: $\hat{x}=N_1 \cdot x  + N_2 $, where $N_1$, $N_2$ are two noise vectors that are sampled from two Gaussian distribution ($N_1  \sim \mathcal{N}(1, \sigma^2_x)$, $N_2  \sim \mathcal{N}(\mu_x, \sigma^2_x)$). 

Existing Sparse Mixture of Experts (SMoE) models provide the same input to the top K-Experts in a TopK setting. In this paper, we propose a novel architecture that enhances model knowledge through Gaussian noise generation, as illustrated in Figure~\ref{fig:S2MoE}. The output of the S2MoE layer is defined by the following equation: 
\begin{equation}
\resizebox{1.0\hsize}{!}{$
f^{\mathrm{S2MoE}}(\boldsymbol{x})=  g\left(x\right) f^{\mathrm{SMoE}}(\boldsymbol{x}) + (1- g\left(x\right))  f^{\mathrm{SMoE}}(\boldsymbol{\hat{x}}),
$
}
\end{equation}

Here, $g(x)$ represents a gating network that combines the SMoE outputs from the original input and the feature-augmented input. The term $1-g(x)$ reflects the model's trade-off between focusing on learning the original features and exploring new ones

\textbf{Learning.}\; Same as ~\cite{fedus_switch_2022}, ~\cite{chi_representation_2022}, we propose the training objective with jointly minimizing the loss of the target task, an auxiliary balancing loss ($\mathcal{L}^{\text {b }}$) and a below uncertainty loss ($\mathcal{L}^{\text {u }}$). For learning under uncertainty, following previous works ~\cite{vo2018composingtextimageimage,Lee_2021_CVPR, chen2024composedimageretrievaltext}, we adopt InfoNCE~\cite{oord2019representationlearningcontrastivepredictive} loss to control the similar between the original input and the noise-augmented input. 
Given $x$, $\hat{x}$ of a mini-batch with $B$ sample as the hidden representations and the noise-augmented one respectively, the uncertainty loss is calculated as follows:

\vspace*{-0.3in}
\begin{equation} \label{eq:nloss}
\mathcal{L}_{\text {u }}\left(x, \hat{x}\right)=\frac{1}{B} \sum_{i=1}^B-\log \frac{\exp \left(\kappa\left(x^i, \hat{x}^i\right)\right)}{\sum_{j=1}^B \exp \left(\kappa\left(x^i, \hat{x}^j\right)\right)} 
\end{equation}
The overall training objective is to minimize:

$
\hspace*{0.4in}
\mathcal{L}=\mathcal{L}_{\text {task }}+\alpha \cdot \mathcal{L}^{\text {b }} +\beta \cdot \mathcal{L}^{\text {u }}
$

where $\alpha$, $\beta$ are coefficients for the balancing loss and uncertainty loss, respectively. The term $\mathcal{L}_{\text {task }}$ is defined by the specific task being learned by the Large Language Models (LLMs), while $\alpha$, $\beta$ are hyperparameters that can be chosen on a case-by-case basis. In practice, we find that $\alpha \approx 0.01$ is an appropriate choice.

\vspace*{-0.05in}


\subsection{S2MoE solves Representation Collapse by Design}
\label{sec:theory}

The Jacobian matrix of S2MoE with respect to $x \in \mathbb{R}^{n \times d}$ is given by:
\vspace*{-0.1in}
\begin{align*}\label{jcob_vqoe2}
\boldsymbol{J_{S2MoE}}=  g\left(x\right) J_{SMoE} + J_{g\left(x\right)}  f_{\mathrm{SMoE}}(x) + 
\\
g\left(x\right)_d N_1^T J_{SMoE}  + (1-J_{g\left(x\right)_d}) N_1^T  f_{\mathrm{SMoE}}(x)
\end{align*}
\begin{equation}\label{jcob_vqoe3}
\Longrightarrow \boldsymbol{J_{S2MoE}}= J_1 + \sum_{j=1}^N c_j e_j^{\top} + \sum_{l=1}^N d_l e_l^{\top} 
\end{equation}
 where $J_1= (J_{g\left(x\right)_d} +(1-J_{g\left(x\right)_d}) N_1^T)  f_{\mathrm{SMoE}}(x)$ ; $c_j=\mathcal{S}(x)_k\left(\delta_{k j}-S_j\right) \boldsymbol{E}(x)_i$ ; $d_l=N_1 c_j$. 
 
 Similar to the Jacobian matrix of SMoE as Section ~\ref{sec:theory_cc}, the Jacobian matrix of S2MoE also consists two terms: (1) $J_1$, which depends on the input token and experts for the final output; and (2) $\sum_{j=1}^{N+N} o_j \boldsymbol{e}_j^{\top}$ indicates to learn better gating function to minimize the task loss. Since $N+N >> N$, this suggests that S2MoE is more effective than SMoE in addressing the representation collapse issue.

\section{Experiments} \label{sec:exp}

We conduct experiments on language model pre-training using various datasets, including Enwik8, Text8~\cite{mahoney_large_2011}, Wikitext-103~\cite{merity_pointer_2017}, and One Billion Words~\cite{chelba2014billionwordbenchmarkmeasuring}. To evaluate performance, we fine-tune the pre-trained models on a range of downstream benchmarks. Additionally, we apply our method to the existing pre-trained language model BERT~\cite{devlin-etal-2019-bert} to demonstrate its effectiveness compared to other SMoE routing methods.

\subsection{Experiment Setting}
Most of our experiments follow the approach of \citet{chen_sparse_2023} and use a base Transformer-XL \cite{dai_transformer_xl_2019} with four decoder layers. 

We compare our S2MoE method with several state-of-the-art routing strategies: (i) \emph{SMoE} \citep{fedus_switch_2022}; (ii) \emph{SMoE-Dropout} \citep{chen_sparse_2023}; (iii) \emph{XMoE} \citep{chi_representation_2022}; and (iv) \emph{StableMoE} \citep{dai_stablemoe_2022}.


\textbf{Pre-training.}\; We train both base and large-scale versions of Transformer-XL on four datasets (Enwik8, Text8, Wikitext-103, and One Billion Words) for 100k iterations, following the implementation in \cite{chen_sparse_2023}. 

\textbf{Fine-tuning.}\; We fine-tune the pre-trained weights for text classification tasks, including \texttt{SST-2} \cite{socher_recursive_2013}, \texttt{SST-5} \cite{socher_recursive_2013}, \texttt{IMDB} \cite{maas_learning_2011}, and \texttt{BANKING77} \cite{casanueva-etal-2020-efficient}. Furthermore, we compare our method with the SMoE baseline using the existing pre-trained language model BERT~\cite{devlin-etal-2019-bert}, following the experimental settings of ~\cite{he-etal-2023-merging}. More implementation details and additional results are provided in the Appendix \ref{sec:appendix}. 

\vspace{-0.1in}
\subsection{Pre-training Result}
\begin{table}[t]
\centering
\caption{Bit-per-character on the \texttt{enwik8}, \texttt{text8}, and Perplexity on the \texttt{WikiText-103}, \texttt{One Billion Words}  test sets, where lower values indicate better performance. Here, $k$ represents the number of experts selected during inference. The best results are highlighted in \textbf{bold}. } \label{tab:pre-train}
\setlength\tabcolsep{4.5pt}
\scriptsize
\begin{tabular*}{0.9\linewidth}{ccccccc}
\toprule
\multicolumn{6}{c}{\texttt{Enwik8}} \\ \midrule
$k$ & SMoE & SMoE-Dropout & StableMoE & XMoE & S2MoE \\ \midrule
1  &    1.22 &    2.47 &     1.22 & 1.23 & \textbf{1.21} \\ 
2  &    1.20 &    1.56 &     1.20 & 1.21 & \textbf{1.19} \\ 
4  &    1.21 &    1.34 &     1.20 & 1.21 & \textbf{1.19} \\ 
8  &    1.21 &    1.27 &     1.21 & 1.21 & \textbf{1.20} \\ 
16 & 1.21 & 1.26  & 1.22 & 1.21 & \textbf{1.21} \\  \midrule
\multicolumn{6}{c}{\texttt{Text8}} \\ \midrule
1  &    1.30 &    2.33 &     1.30 & 1.30 & \textbf{1.29} \\
2  &    1.29 &    1.56 &     1.28 & 1.28 & \textbf{1.27} \\
4  &    1.29 &    1.40 &      1.29 & 1.28 & \textbf{1.28} \\
8  &    1.29 &    1.34 &      1.29 & 1.29 & \textbf{1.28} \\
16 & 1.29 & 1.33 &  1.30 & 1.29 & \textbf{1.29} \\ \midrule
\multicolumn{6}{c}{\texttt{WikiText-103}} \\ \midrule
1  &    32.38 &    130.11 &     31.88 & 32.83 & \textbf{31.31} \\
2  &    30.16 &    58.37 &     29.97 & 30.34 & \textbf{29.63} \\
4  &    30.34 &    41.88 &      30.34 & 30.71 & \textbf{29.97} \\
8  &    30.94 &    36.93 &      31.12 & 31.21 & \textbf{30.59} \\
16 & 31.31 & 37.82  &  31.81 & 31.48 & \textbf{31.09} \\ \midrule
\multicolumn{6}{c}{\texttt{One Billion Word}} \\ \midrule
1  &    61.82 &    197.56 &     61.89 & 62.04 & \textbf{60.79} \\
2  &    58.00 &    93.19 &     58.25 & 58.33 & \textbf{57.44} \\
4  &    58.68 &    69.90 &      58.77 & 58.99 & \textbf{57.94} \\
8  &    59.79 &    62.26 &      59.77 & 59.88 & \textbf{58.81} \\
16 & 60.38 & 61.67  &  60.44 & 60.29 & \textbf{59.39} \\
\bottomrule
\end{tabular*}
\vspace*{-0.1in}
\end{table}

{\textbf{Base training.}\;}Table~\ref{tab:pre-train} presents the pre-training results for four datasets (\texttt{enwik8}, \texttt{text8}, \texttt{WikiText-103}, and \texttt{One Billion Words}). We observe that S2MoE significantly outperforms the baseline SMoE, as well as advanced routing methods such as XMoE \cite{chi_representation_2022} and StableMoE \cite{dai_stablemoe_2022} on the four all pre-training datasets. The advantage of S2MoE training lies in its inference efficiency by using fewer experts. Notably, S2MoE significantly outperforms SMoE on \texttt{text8} when using only one expert. It also surpasses SMoE-Dropout (two experts) on \texttt{WikiText-103}, reducing perplexity from \textbf{93.19} to \textbf{60.79} with just one expert. When S2MoE uses only one expert, it \textbf{reduces FLOPs by 28\%} compared to methods like SMoE and SMoE-Dropout, which use two experts, while maintaining competitive performance.

{\textbf{Large training.}\;} . Table~\ref{tab:pre-train-large} reports the BPC on the enwik8 dataset using large Transformer-XL. We observe the gap between our S2MoE and the baselines becomes more significant, indicating our S2MoE enjoys good scalability with the model complexity. S2MoE consistently outperforms both baselines, regardless of backbone size or the number of experts activated, demonstrating its potential to scale up effectively in large language models.

\subsection{Fine-tuning Result}

\begin{table}[t] 
\resizebox{\linewidth}{!}{%
\begin{tabular}{@{}lccccccc@{}}
\toprule
Model        &  \multicolumn{4}{c}{Transformer-XL} & \multicolumn{3}{c}{BERT} \\ \midrule
Method        &  {SST-2} & {SST-5} & {IMDB} & {BANKING77} & {MRPC} & {QNLI} & {SST-2} \\ \midrule
S2MoE     & \textbf{83.6}             & \textbf{41.4}             & \textbf{89.5}            & \textbf{87.2}   & \textbf{78.7} & \textbf{90.6} & \textbf{92.8}            \\
SMoE        & 80.8                      & 40.4                      & 88.6                     & 80.2  & 74.5 & 90.0 & 92.2                      \\
SMoE-Dropout    & 81.8                      & 40.0                      & 89.1                     & 77.3  & - & - & -                     \\
XMoE        & 81.3                      & 40.3                      & 88.7                     & 82.7 & - & - &                        \\
StableMoE        & 82.5                      & 41.1                      & 88.5                     & 78.6 & - & - & -                     \\
MEO~\cite{he-etal-2023-merging}        & -                      & -                    & -                     & - & 76.2 & 90.4 & 92.3                     \\ \bottomrule
\end{tabular}}
\caption{Accuracy of the model after fine-tuned on various datasets. Higher is better, best results are in bold.} \label{tab:finetune}
\vspace*{-0.1in}
\end{table}

\textbf{Pre-training weights.}\; We report the results of the fine-tuning experiment on the \texttt{SST-2}, \texttt{SST-5}, \texttt{IMDB}, and \texttt{BANKING77} datasets in Table ~\ref{tab:finetune}, using Transformer-XL pre-trained on \texttt{enwik8}.  Overall, S2MoE consistently achieves higher accuracy compared to other baselines across all datasets. 

\textbf{BERT.}\;We implement Sparse Mixture of Experts for BERT~\cite{devlin-etal-2019-bert}, following the MEO approach ~\cite{he-etal-2023-merging}. We present the fine-tuning results on the \texttt{MRPC}~\cite{dolan-brockett-2005-automatically}, \texttt{QNLI}~\cite{wang-etal-2018-glue}, and \texttt{SST-2} datasets using S2MoE, comparing it with SMoE and the MEO baseline in Table ~\ref{tab:finetune}
. The results demonstrate that our method is not only effective for pre-training tasks but also performs effectively on existing pre-trained models, such as those in the BERT family.

\section{Conclusion}
\vspace*{-0.1in}
In this research, we explored the potentials and limitations of SMoE for training large language models (LLMs) and introduced Uncertain Sparse Mixture of Experts (S2MoE) to enhance expert learning capacity while mitigating the collapse issue among experts. As a result, S2MoE is able to learn more robust expert representations while addressing the representation collapse commonly seen in conventional SMoE training.Experiments on both pre-training and fine-tuning tasks demonstrated that S2MoE enables more efficient and effective training and inference compared to advanced routing strategies.

\section*{Limitations}
Our study centers on improving the efficiency and effectiveness of training large language models (LLMs) using SMoE. While the results are promising, our experiments were limited to medium-scale datasets and a base Transformer-XL model due to computational constraints. Therefore, further empirical evaluations are necessary to validate the scalability of S2MoE and other SMoE strategies on modern LLMs and larger datasets. 

\section*{Ethics Statement}
Despite promising results, training large-scale LLMs remains inherently costly and demands significant computational resources, which must be carefully managed. Additionally, our paper utilized web-sourced data, which is known to contain gender and racial biases, necessitating further efforts to mitigate these negative impacts. Lastly, while our study marks a promising step toward advancing the development of new LLMs, it underscores the need for careful regularization to prevent potential misuse in harmful applications.

\bibliography{acl_latex}

\appendix 

\section{Appendix}
\label{sec:appendix}
\subsection{Implementation Details}

The base Transformer-XL variant~\citep{chen_sparse_2023} comprises four Transformer decoder layers, each with an input dimension of 256. Each layer includes a self-attention mechanism with eight attention heads, followed by a feedforward neural network (FFN) that has an inner dimension of 512. The dropout ratio is set at 0.1. We divide the FFN into 16 experts, each with the same dimensions. For the larger variants, we scale the model up to twelve layers.

Our experiments are based on the publicly available SMoE-Dropout implementation \citep{chen_sparse_2023}\footnote{\url{https://github.com/VITA-Group/Random-MoE-as-Dropout}}. The pre-training experiments were conducted using a single H100 GPU, while the fine-tuning experiments were performed on a single A100 GPU. It is important to note that parallel training on multiple GPUs may produce different results.

\subsection{Pre-training Experiments}
We provide the S2MoE implementation details for pre-training our Transformer-XL base and large on \texttt{enwik8}, \texttt{text8}, \texttt{WikiText-103}, and \texttt{One Billion Word} in Table \ref{tab:A1}.

\begin{table}[!ht]
\centering
\caption{Implementation details for pre-training experimentson \texttt{enwik8}, \texttt{text8}, \texttt{WikiText-103}, and \texttt{One Billion Word} datasets. }
\scriptsize
\setlength\tabcolsep{3.06pt}
\label{tab:A1}
\begin{tabular}{lccccc}
\midrule
Dataset   & Input length & Batch size & Optimizer & Lr   & \# Iterations \\ \midrule
\texttt{enwik8}      & 512          & 48          & Adam      & 2.5e-4 & 100k         \\
\texttt{text8} & 512          & 48         & Adam      & 2.5e-4 & 100k         \\ 
\texttt{WikiText-103} & 512          & 22         & Adam      & 2.5e-4 & 100k         \\ 
\texttt{One Billion Word} & 512          & 11         & Adam      & 2.5e-4 & 100k         \\ \midrule
\end{tabular}
\end{table}


\subsection{Fine-tuning Experiments}
\noindent To perform the fine-tuning experiments, we utilize the same model architecture as in the pre-training phase. Table \ref{tab:A2} presents the implementation details for the fine-tuning experiments conducted across four different datasets. 

\begin{table}[!ht]
\centering
\caption{Implementation for fine-tuning experiments on downstream tasks. }
\scriptsize
\setlength\tabcolsep{4.86pt}
\label{tab:A2}
\begin{tabular}{lccccc}
\midrule
Dataset   & Input length & Batch size & Optimizer & Lr   & \# Epochs \\ \midrule
\texttt{SST-2}     & 512          & 16         & Adam      & 1e-4 & 15         \\
\texttt{SST-5}     & 512          & 16         & Adam      & 1e-4 & 15         \\
\texttt{IMDB}      & 512          & 4          & Adam      & 1e-4 & 15         \\
\texttt{BANKING77} & 512          & 16         & Adam      & 1e-4 & 15         \\ \midrule
\end{tabular}
\end{table}
\subsection{Additional Results}

We trained a large Transformer-XL model with 12 decoder layers and 64 experts. The results are reported as Table~\ref{tab:pre-train-large}.
\begin{table}[t]
\centering
\caption{Perplexity on the \texttt{Wikitext-103} test set using the Transformer-XL large models. Lower is better. The best results are highlighted in \textbf{bold}. } \label{tab:pre-train-large}
\setlength\tabcolsep{4.5pt}
\scriptsize
\begin{tabular*}{0.8\linewidth}{cccccc}
\toprule
\multicolumn{5}{c}{\texttt{Large Transformer-XL}} \\ \midrule
$k$ &  SMoE-Dropout & StableMoE & XMoE & S2MoE \\ \midrule
1  &       84.86 &     26.80 & 27.26 & \textbf{26.20} \\ 
2  &       40.02 &     24.24 & 24.28 & \textbf{24.12} \\ 
4  &      31.41 &     24.56 & 24.54 & \textbf{24.24} \\ 
8  &       28.64 &     25.67 & 25.05 & \textbf{24.90} \\ 
16 &  28.30  & 26.85 & \textbf{25.54} & 26.00 \\ 
\bottomrule
\end{tabular*}
\vspace*{-0.1in}
\end{table}

One of the key hyperparameters of the S2MoE method is 
$\beta$, which determines the quality of feature generation from Gaussian noise. The hyperparameter $\beta$ can be learned from data. In practice, we have found that values of $\beta$ in the range of $(0.1, 0.01)$ are effective as Table ~\ref{tab:tuning}.
\begin{table}[!ht]
\centering
\caption{Tuning $\beta$ on \texttt{enwik8} dataset. }
\scriptsize
\setlength\tabcolsep{4.86pt}
\label{tab:tuning}
\begin{tabular}{lc}
\midrule
$\beta$   & Transformer-XL  \\ \midrule
\texttt{1.0}     & 1.202             \\
\texttt{0.5}     & 1.186          \\
\texttt{0.1}     & 1.159             \\
\texttt{0.01}    & 1.163            \\ \midrule
\end{tabular}
\end{table}

\subsection{Representation Collapse in SMoE}
\label{sec:theory_cc}

Following ~\citep{chi_representation_2022} and ~\citep{do2023hyperrouterefficienttraininginference}, we illustrate the representation collapse issue using the Jacobian matrix approach. Specifically, the Jacobian matrix of the SMoE with respect to $x \in \mathbb{R}^{n \times d}$ is given as:
\vspace*{-0.1in}
\begin{equation}\label{jcob_smoe_p1}\resizebox{1.0\hsize}{!}{$
\boldsymbol{J_{SMoE}}=\mathcal{S}(x)_k \boldsymbol{J}^{\mathrm{FFN}} + \sum_{j=1}^N \mathcal{S}(x)_k\left(\delta_{k j}-S_j\right) \boldsymbol{E}(x)_i \boldsymbol{e}_j^{\top} \notag
$}
\end{equation}
\vspace*{-0.3in}
\begin{equation}\label{jcob_smoe_p2}
\Longrightarrow \boldsymbol{J_{SMoE}}= \mathcal{S}(x)_k \boldsymbol{J}^{\mathrm{FFN}} + \sum_{j=1}^N \boldsymbol{c}_j \boldsymbol{e}_j^{\top},
\end{equation}
\vspace*{-0.01in}
where $\boldsymbol{c}_j=\mathcal{S}(x)_k\left(\delta_{k j}-S_j\right) \boldsymbol{E}(x)_i$. The fist part of Equation ~\ref{jcob_smoe_p2}, $\mathcal{S}(x)_k \boldsymbol{J}^{\mathrm{FFN}}$, represents the contribution from the input token and experts to the final output. The second part, (2) $\sum_{j=1}^N \boldsymbol{c}_j \boldsymbol{e}_j^{\top}$ relates to learning an improved gating function to minimize task loss. Furthermore, Equation ~\ref{jcob_smoe_p2}  is recommended to be updated as a linear combination of expert embeddings. Due to $N << d$ in practice, the above equation illustrates representation collapse from $\mathbb{R}^d$ to $\mathbb{R}^N$. 

\end{document}